\documentclass[letterpaper]{article}
\usepackage{ijcai13}
\usepackage{times}
\usepackage{latexsym}
\usepackage{graphicx}
\usepackage{color}
\usepackage{multirow}
\usepackage{amsmath}
\usepackage{amssymb}
\usepackage{psfig}
\usepackage{subfig}
\usepackage[ruled,vlined,linesnumbered]{algorithm2e}

\newcommand{\OPEN} {{\textsc{Open}}}
\newcommand{\CLOSED} {{\textsc{Closed}}}
\newcommand{\astar} {\ensuremath{A^\ast}}
\newcommand{\lazyastar} {\ensuremath{LA^\ast}}
\newcommand{\rationallazyastar} {\ensuremath{RLA^\ast}}
\newcommand{\optcost} {\ensuremath{C^\ast}}
\newcommand{\astarmax}{\ensuremath{\astar_{MAX}}}

\newcommand{\citet}[1]{\citeauthor{#1} (\citeyear{#1})}

\newcommand{\comment}[1]{}

\title{Towards Rational Deployment of Multiple Heuristics in A*}

\author{David Tolpin \\ \begin{Large} {\bf Tal Beja} \end{Large} \\ \begin{Large}  {\bf Solomon Eyal Shimony} \end{Large}  \\
CS Department \\
Ben-Gurion University \\
Israel\\
 \{tolpin,bejat,shimony\}@cs.bgu.ac.il
\And
Ariel Felner \\ ISE Department \\
Ben-Gurion University \\
Israel\\
felner@bgu.ac.il
\And
Erez Karpas \\
Faculty of IE\&M \\
Technion\\
Israel\\
karpase@gmail.com
}

\begin{document}

\maketitle

\begin{small}
\begin{abstract}
The obvious way to use several admissible heuristics in \astar~is to take their
maximum.
In this paper we aim to reduce the time spent on computing heuristics.
We discuss {\em Lazy \astar},  a variant of \astar~where heuristics
are evaluated lazily: only when they are essential to a decision to be
made in the \astar~search process.  We present a new rational
meta-reasoning based scheme, {\em rational lazy \astar}, which
decides whether to compute the more expensive heuristics at all, based on a
myopic value of information estimate. Both methods are examined theoretically.
Empirical evaluation on several domains supports the theoretical results, and
shows that lazy \astar~and rational lazy \astar~are state-of-the-art heuristic
combination methods.
\end{abstract}
\end{small}


\vspace{-0.4cm}
\section{Introduction}

The \astar~algorithm \cite{ASTR68} is a
best-first heuristic search algorithm guided by the cost function $f(n)=g(n)+h(n)$. 
If the heuristic $h(n)$ is admissible (never overestimates the real cost to the goal)
then the set of nodes expanded by \astar~is both necessary and sufficient to find the optimal path to the goal~\cite{ASTR85}.

This paper examines the case where we have several available admissible
heuristics. Clearly, we can evaluate all these heuristics, and use their {\em
maximum} as an admissible heuristic, a scheme we call \astarmax.
The problem with naive maximization is that all the heuristics are
computed for all the generated nodes.
In order to reduce the time spent on heuristic computations, Lazy $\astar$ (or
\lazyastar, for short) evaluates the heuristics one at a time, lazily. When a node $n$ is
generated, \lazyastar~only computes one heuristic, $h_1(n)$, and adds $n$ to
\OPEN.
Only when $n$ re-emerges as the top of \OPEN~is another heuristic, $h_2(n)$,
evaluated; if this results in an increased heuristic estimate, $n$ is
re-inserted into \OPEN.
This idea was briefly mentioned by~\citet{zhang-bacchus:aaai-2012} in the
context of the MAXSAT heuristic for planning domains.
\lazyastar~is as informative as \astarmax, but can significantly reduce
search time, as we will not need to compute $h_2$ for many nodes.
In this paper we provide a deeper
examination of \lazyastar, and characterize the savings that it can lead to.
In addition, we describe several technical optmizations for \lazyastar.

\lazyastar~reduces the search time, while maintaining the
informativeness of \astarmax.
However, as noted by \citet{domshlak-et-al:jair-2012},  if the goal is to reduce
search time, it may be better to compute a fast heuristic on
several nodes, rather than to compute a slow but informative heuristic on only one
node.
Based on this idea, they formulated {\em selective max} (Sel-MAX), an online
learning scheme which chooses one heuristic to compute at each
state. Sel-MAX chooses to compute the more expensive heuristic $h_2$ for
node $n$ when its classifier predicts that $h_2(n) - h_1(n)$ is greater than
some threshold, which is a function of heuristic computation times and the average
branching factor.
\citet{INCJUR} showed that randomizing a heuristic and applying {\em
bidirectional pathmax} (BPMX) might sometimes be faster than evaluating all
heuristics and taking the maximum. This technique is only useful in undirected
graphs, and is therefore not applicable to some of the domains in this paper.
Both Sel-MAX and Random compute the resulting heuristic {\em once},
before each node is added to \OPEN~while \lazyastar~computes the heuristic
lazily, in different steps of the search. In addition, both randomization and Sel-MAX save heuristic computations and thus reduce search time in many cases.
However, they  might be less informed than pure maximization and as a result
expand a larger number of nodes.


We then combine the ideas of lazy heuristic evaluation and of trading off more node expansions
for less heuristic computation time, into a {\em new} variant of \lazyastar~called
{\em rational lazy} \astar~(\rationallazyastar).
\rationallazyastar~is based on rational meta-reasoning, and uses a myopic {\em
value-of-information} criterion to decide whether to compute $h_2(n)$ or to
bypass the computation of $h_2$ and expand $n$ immediately when $n$ re-emerges
from \OPEN. \rationallazyastar~aims to reduce search time, even at the expense of more node expansions than \astarmax.

Empirical results on variants of the 15-puzzle and on numerous planning
domains demonstrate that \lazyastar~and \rationallazyastar~lead
to state-of-the-art performance in many cases.

\vspace{-0.2cm}
\section{Lazy \astar}

Throughout this paper we assume for clarity that we have two available admissible heuristics, $h_1$ and $h_2$.
Extension to multiple heuristics is straightforward, at least for \lazyastar.
Unless stated otherwise, we assume that $h_1$ is faster to compute than $h_2$
but that $h_2$ is {\em weakly more informed}, i.e., $h_1(n) \leq h_2(n)$ for
the majority of the nodes $n$, although counter cases where $h_1(n) > h_2(n)$
are possible.
We say that $h_2$ {\em dominates} $h_1$, if such counter cases do not
exist and $h_2(n) \geq h_1(n)$ for {\em all} nodes $n$.
We use $f_1(n)$ to denote $g(n)+h_1(n)$. Likewise, $f_2(n)$
denotes $g(n)+h_2(n)$, and $f_{max}(n)$ denotes $g(n) + \max(h_1(n),h_2(n))$.
We denote the cost of the optimal solution by $\optcost$. Additionally, we
denote the computation time of $h_1$ and of $h_2$ by $t_1$ and $t_2$,
respectively and denote the overhead of an {\em insert/pop} operation in
\OPEN~by $t_o$. Unless stated otherwise we assume that $t_2$ is much greater
than $t_1 + t_o$. \lazyastar~thus mainly aims to reduce computations of $h_2$.

\begin{algorithm}[t]
\begin{small}
\KwIn{LAZY-\astar}
    Apply all heuristics to Start\\
    Insert Start into \OPEN \\
    \While{\OPEN~ not empty}{
        $n$ $\gets$ best node from \OPEN \\
        \If{Goal(n)}{
            \Return trace(n)\\
        }
        \If{$h_2$ was not applied to $n$}{
            Apply $h_2$ to $n$ \\
            insert $n$ into \OPEN\\
            continue ~~~~~~ //next node in OPEN\\
        }
        \ForEach{child $c$ of $n$}{
            Apply $h_1$ to $c$.\\
            insert $c$ into \OPEN\\
        }
        Insert $n$ into \CLOSED\\
    }
    \Return FAILURE
\caption{Lazy \astar}
\label{LAZYA}
\end{small}
\end{algorithm}

The pseudo-code for  \lazyastar~is depicted as Algorithm \ref{LAZYA}, and is
very similar to \astar.
In fact, without lines 7 -- 10, \lazyastar~would be identical to
\astar~using the $h_1$ heuristic.
When a node $n$ is generated we only compute $h_1(n)$ and $n$ is added to
\OPEN ~(Lines 11 -- 13), without computing $h_2(n)$ yet.
When $n$ is first removed from \OPEN~(Lines 7 -- 10), we compute $h_2(n)$ and
reinsert it into \OPEN, this time with $f_{max}(n)$.

It is easy to see that \lazyastar~is as informative as \astarmax, in
the sense that both \astarmax~and \lazyastar expand a node $n$ only
if $f_{max}(n)$ is the best $f$-value in \OPEN.  Therefore,
\lazyastar~and \astarmax~generate and expand and the same set of
nodes, up to differences caused by tie-breaking.


In its general form \astar~generates many nodes that it does not expand. These
nodes, called {\em surplus} nodes~\cite{Felner2012}, are in \OPEN~when we
expand the goal node with $f=\optcost$. All nodes in \OPEN~with $f>\optcost$ are
surely surplus but some nodes with $f=\optcost$ may also be surplus. The number
of surplus nodes in OPEN can grow exponentially in the size of the domain, resulting in
significant costs.

\lazyastar~avoids $h_2$ computations for many of these surplus nodes. Consider
a node $n$ that is generated with $f_1(n) > \optcost$. This node is inserted
into \OPEN~but will never reach the top of \OPEN, as the goal node will be found
with $f=\optcost$. In fact, if \OPEN~breaks ties in favor of small $h$-values,
the goal node with $f=\optcost$ will be expanded as soon as it is generated and such
savings of $h_2$ will be obtained for some nodes with $f_1=\optcost$ too. We
refer to such nodes where we saved the computation of $h_2$ as {\em good} nodes. Other nodes,
those with $f_1(n) < \optcost$ (and some with $f_1(n) = \optcost$) are called
{\em regular nodes} as we apply both heuristics to them.

\astarmax~computes both $h_1$ and $h_2$ for all generated nodes, spending time
$t_1 + t_2$ on all generated nodes. By contrast, for {\em good} nodes
\lazyastar~only spends $t_1$, and saves $t_2$. In the basic implementation of
\lazyastar~(as in algorithm \ref{LAZYA}) {\em regular} nodes are inserted into
OPEN twice, first for $h_1$ (Line 13) and then for $h_2$ (Line 9) while {\em good} nodes only enter \OPEN~once (Line 13). Thus, \lazyastar~has some extra overhead
of \OPEN~operations for {\em regular nodes}. We distinguish between
3 classes of nodes:\\ {\bf (1)} {\em expanded regular} (ER) --- nodes that
were expanded after both heuristics were computed.\\ {\bf (2)} {\em surplus regular} (SR) --- nodes
for which $h_2$ was computed but are still in \OPEN~when the goal was found.\\ {\bf
(3)} {\em surplus good} (SG) --- nodes for which only $h_1$ was computed by \lazyastar~when
the goal was found.

\begin{table}[tbh]\vspace{-0.3cm}
\begin{center}
\begin{tabular}{|c|c|c|c|}
\hline
Alg & ER & SR & SG \\
\hline
\astarmax & $t_1 + {\bf t_2} + 2t_o$ & $t_1+ {\bf t_2} + t_o$ & $t_1 + {\bf t_2} + t_o$ \\
\lazyastar & $t_1 + {\bf t_2} + 4t_o$ & $t_1+ {\bf t_2} + 3t_o$ & $t_1 + t_o$ \\
\hline
\end{tabular}
\end{center}\vspace{-0.4cm}
\caption{Time overhead for \astarmax~and for $\lazyastar$}\vspace{-0.3cm}
\label{TIME}
\end{table}

The time overhead of \astarmax~and \lazyastar~is summarized in
Table~\ref{TIME}. \lazyastar~incurs more \OPEN~operations overhead, but saves
$h_2$ computations for the SG nodes.
When $t_2$ ({\bf boldface} in table \ref{TIME}) is significantly greater than
both $t_1$ and $t_o$  there is a clear advantage for \lazyastar, as seen in
the SG column.

\section{Enhancements to Lazy $\astar$}

Several enhancements can improve basic \lazyastar~(Algorithm
\ref{LAZYA}), which are effective especially if $t_1$ and $t_o$ are not negligible.

\subsection{\OPEN~bypassing}

Suppose node $n$ was just generated, and let $f_{best}$
denote the best $f$-value currently in \OPEN.
\lazyastar~evaluates $h_1(n)$ and then inserts $n$ into \OPEN. However, if
$f_1(n) \leq f_{best}$, then  $n$ will immediately reach the
top of \OPEN ~and $h_2$ will be computed.
In such cases 
we can choose to compute $h_2(n)$ right away (after Line 12 in Algorithm~\ref{LAZYA}), thus saving the overhead of
inserting $n$ into \OPEN ~and popping it again at the next step ($= 2 \times
t_o$).
For such nodes, \lazyastar~is identical to $\astar_{MAX}$, as both heuristics
are computed before the node is added to \OPEN.
This enhancement is called {\em OPEN bypassing} (OB). It is a reminiscent of the {\em immediate expand} technique applied to generated nodes~\cite{DBLP:conf/aaai/SternKFH10,AAMAS2}. The same
technique can be applied when $n$ again reaches the top of \OPEN~when evaluating $h_2(n)$ ; if $f_2(n) \leq f_{best}$, expand $n$ right away.
With OB, \lazyastar~will incur the extra overhead of two
\OPEN~cycles only for nodes $n$ where $f_1(n) > f_{best}$ and then later
$f_2(n) > f_{best}$.

\subsection{Heuristic bypassing}

{\em Heuristic bypassing} (HBP) is a technique that allows \astarmax~to omit
evaluating one of the two heuristics.
HBP is probably used by many implementers, although to the best of our
knowledge, it never appeared in the literature.
HBP works for a node $n$ under the following two preconditions: {\bf (1)} the
operator between $n$ and its parent $p$ is bidirectional, and {\bf (2)} both
heuristics are {\em consistent}~\cite{INCJUR}.


\begin{figure}\vspace{-0.3cm}
    \begin{centering}
    \includegraphics[height=1.2cm]{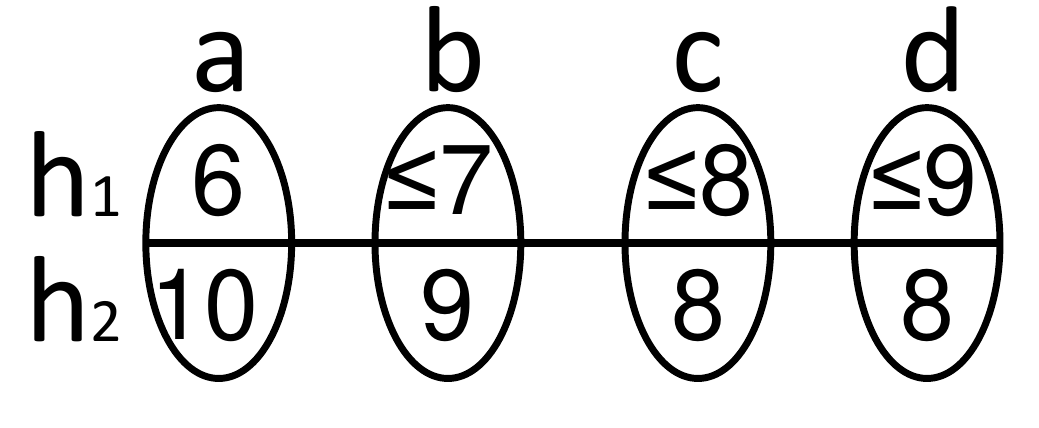}
     \begin{small}\vspace{-0.4cm}
	\caption{Example of HBP}\vspace{-0.3cm}
	\label{fig:bypassing}
\end{small}
\end{centering}
\end{figure}

Let $C$ be the cost of the operator. Since the heuristic is consistent we
know that $|h(p)-h(n)| \leq C$. Therefore, $h(p)$ provides the following upper-
and lower-bounds on $h(n)$ of $h(p) - C \leq h(n) \leq h(p) + C$.
We thus denote $\underline{h(n)}=h(p)-C$ and $\overline{ h(n)}=h(p)+C$.

To exploit HBP in \astarmax, we simply skip the computation of $h_1(n)$
if $\overline{h_1(n)} \leq \underline{h_2(n)}$, and vice versa.
For example, consider node $a$ in
Figure~\ref{fig:bypassing}, where all operators cost 1, $h_1(a)=6$, and
$h_2(a)=10$.
Based on our bounds $h_1(b) \leq 7$ and $h_2(c) \geq 9$.
Thus, there is no need to check $h_{1}(b)$ as $h_2(b)$ will surely be the
maximum.
We can propagate these bounds further to node $c$. $h_2(c)=8$ while $h_1(c)
\leq 8$ and again there is no need to evaluate $h_1(c)$.
Only in the last node $d$ we get that  $h_2(d)=8$ but since $h_1(c) \leq 9$
then  $h_1(c)$ can potentially return the maximum and should thus be evaluated.


HBP can be combined in \lazyastar~in a number of ways. We describe the variant
we used. \lazyastar~aims to avoid needless computations of $h_2$. Thus, when
$\overline{h_1(n)} < \underline{h_2(n)}$, we delay the computation of
$h_2(n)$ and add $n$ to \OPEN~ with $f(n)=g(n)+ \underline{h_2(n)}$ and
continue as in \lazyastar. In this case, we saved $t_1$, delayed $t_2$ and used
$\underline{h_2(n)}$ which is more informative than $h_1(n)$.
If, however, $\overline{h_1(n)} \geq \underline{h_2(n)}$, then we compute
$h_1(n)$ and continue regularly.
We note that HBP incurs the time and memory overheads of computing and storing
four bounds and should only be applied if there is enough memory and if $t_1$
and especially $t_2$ are very large.





\vspace{-0.1cm}
\section{Rational Lazy $\astar$}\vspace{-0.1cm}


\lazyastar~offers us a very strong guarantee, of expanding the same set of nodes
as \astarmax. However, often we would prefer to expand more states, if it means
reducing search time. We now present \textit{Rational Lazy A*}
(\rationallazyastar), an algorithm which attempts to
optimally manage this tradeoff.

Using principles of rational meta-reasoning
\cite{RussellWefald}, theoretically every algorithm action (heuristic function evaluation, node
expansion, open list operation) should be treated as an action in a sequential
decision-making meta-level problem: actions should be chosen so as to
achieve the minimal expected search time. However, the appropriate
general meta-reasoning problem is extremely hard to define precisely and
to solve optimally.

Therefore, we focus on just one decision type,
made in the context of \lazyastar, when $n$ re-emerges from \OPEN~(Line 7).
We have two options: {\bf (1)}  Evaluate the second heuristic $h_2(n)$ and add
the node back to \OPEN~(Lines 7-10) like \lazyastar, or {\bf (2)} bypass the
computation of $h_2(n)$ and expand $n$ right way (Lines 11 -13), thereby
saving time by not computing $h_2$, at the risk of additional expansions and evaluations of $h_1$.
In order to choose rationally, we define a criterion based on value of
information (VOI) of evaluating $h_2(n)$ in this context.


The only addition of \rationallazyastar~to \lazyastar~is the option to bypass
$h_2$ computations (Lines 7-10).
Suppose that we choose to compute $h_2$ --- this results in one of the
following outcomes:\\
~~{\bf 1:} $n$ is still expanded, either now or eventually.\\
~~{\bf 2:} $n$ is re-inserted into \OPEN, and the goal is found without ever expanding $n$.

Computing $h_2$ is helpful only in outcome 2, where
potential time savings are due to pruning a search subtree at the expense of
the time $t_2(n)$. However, whether outcome 2 takes place after a given state
is not known to the algorithm until the goal is found, and the algorithm must
decide whether to evaluate $h_2$ according to what it \textit{believes to be}
the probability of each of the outcomes. We derive a \textit{rational policy}
for when to evaluate $h_2$, under the myopic assumption that the algorithm
continues to behave like \lazyastar~afterwards (i.e., it will never again
consider bypassing the computation of $h_2$).

The time wasted by being sub-optimal in deciding whether to evaluate $h_2$ is
called the {\em regret} of the decision. If $h_2(n)$ is not helpful and we decide
to compute it, the effort for evaluating $h_2(n)$ turns out to be wasted. On the
other hand, if $h_2(n)$ is helpful but we decide to bypass it, we needlessly expand $n$.
Due to the myopic assumption, \rationallazyastar~would evaluate both $h_1$ and $h_2$ for all successors of
$n$.

\begin{table}[h]\vspace{-0.2cm}
\begin{small}
\begin{center}
\begin{tabular}{|l|c|c|}
\hline
               & Compute $h_2$ & Bypass $h_2$\\
\hline
$h_2$ helpful &   0            & $t_e+(b(n)-1)t_d$\\
\hline
$h_2$ not helpful & $t_d$      & 0 \\
\hline
\end{tabular}
\end{center}
\end{small}\vspace{-0.4cm}
\caption{Regret in Rational Lazy A*}
\label{tbl:rational-lazy-a-time}
\end{table}\vspace{-0.2cm}

Table~\ref{tbl:rational-lazy-a-time}
summarizes the regret of each possible decision, for each possible future
outcome; each column in the table represents a decision, while each row
represents a future outcome.
In the table, $t_d$ is the to time compute $h_2$ and re-insert $n$ into
\OPEN~thus delaying the expansion of $n$, $t_e$ is the time to remove $n$ from \OPEN,
expand $n$, evaluate $h_1$ on each of the $b(n)$ (``local branching factor'')
children $\{n'\}$ of $n$, and insert $\{n'\}$ into the open list.
Computing $h_2$ needlessly wastes time $t_d$.
Bypassing $h_2$ computation when $h_2$ would have been helpful wastes
$t_e+b(n)t_d$ time, but because computing $h_2$ would have cost $t_d$, the
regret is $t_e+(b(n)-1)t_d$.

Let us denote the probability that $h_2$ is helpful by
$p_h$.
The expected regret of computing $h_2$ is thus $(1-p_h) t_d$.
On the other hand, the expected regret of bypassing $h_2$ is $p_h(
t_e+(b(n)-1)t_d)$. As we wish to minimize the expected regret, we should thus evaluate $h_2$ just when:
\begin{equation}\vspace{-0.1cm}
(1-p_h) t_d < p_h (t_e+(b(n)-1)t_d)
\end{equation}
or equivalently:
\begin{equation}\vspace{-0.1cm}
(1-b(n) p_h) t_d < p_h t_e 
\end{equation}

If $p_h b(n) \ge 1$, then the expected regret is minimized by always
evaluating $h_2$, regardless of the values of $t_d$ and $t_e$.
In these cases, \rationallazyastar~cannot be expected to do better than
\lazyastar.
For example, in the 15-puzzle and its variants, the
effective branching factor is $\approx 2$. Therefore, if $h_2$ is expected to be helpful for more than half of the nodes $n$
on which \lazyastar~evaluates $h_2(n)$, then one should simply use \lazyastar.

\begin{table*}\vspace{-0.7cm}
\begin{centering}
\begin{small}
\begin{tabular}{|c|| r r || r r r r || r r r r r | } \hline
 &\multicolumn{2}{|c||}{\astar}&\multicolumn{4}{c||}{\lazyastar }&\multicolumn{5}{c|}{\rationallazyastar (Using Eq. \ref{eqn:criterion-gamma})} \\
\hline
lookahead & generated & time & generated & Good1 & $h_2$ & time & generated & Good1   & Good2  & $h_2$     & time \\ \hline
2 & 1,206,535 & 0.707 & 1,206,535 & 391,313 & 815,213 & 0.820 & 1,309,574 & 475,389 & 394,863 & 439,314 & 0.842 \\ \hline
4 & 1,066,851 & 0.634 & 1,066,851 & 333,047 & 733,794 & 0.667 & 1,169,020 & 411,234 & 377,019 & 380,760 & 0.650 \\ \hline
6 & 889,847 & {\bf 0.588} & 889,847 & 257,506 & 632,332 & 0.533 & 944,750 & 299,470 & 239,320 & 405,951 & 0.464 \\ \hline
8 & 740,464 & 0.648 & 740,464 & 196,952 & 543,502 & {\bf 0.527} & 793,126 & 233,370 & 218,273 & 341,476 & 0.377 \\ \hline
10 & 611,975 & 0.843 & 611,975 & 145,638 & 466,327 & 0.671 & 889,220 & 308,426 & 445,846 & 134,943 & {\bf 0.371} \\ \hline
12 & {\bf 454,130} & 0.927 & {\bf 454,130} & 95,068 & 359,053 & 0.769 & 807,846 & 277,778 & 428,686 & {\bf 101,378} & 0.429 \\ \hline
\end{tabular}
\end{small}
\end{centering}\vspace{-0.2cm}
\caption{Weighted 15 puzzle: comparison of $A^*_{\max}$, Lazy $A^*$, and Rational Lazy $A^*$}\vspace{-0.2cm}
\label{tbl:rational-lazy-a-star}
\end{table*}

For $p_h b(n) < 1$,  the decision of whether to evaluate $h_2$
depends on the values of $t_d$ and $t_e$:
\begin{equation}\vspace{-0.2cm}
\mbox{\bf evaluate }h_2\mbox{ \bf if }t_d<\frac {p_h} {1-p_hb(n)} t_e
\label{eqn:criterion-general}
\end{equation}
Denote by $t_c$ the time to generate the children of $n$. Then:
\begin{align}
t_d&=t_2+t_o\nonumber\\
t_e&=t_o + t_c+b (n) t_1 + b(n) t_o
\label{eqn:t-del-exp-expanded}
\end{align}
By substituting
(\ref{eqn:t-del-exp-expanded}) into (\ref{eqn:criterion-general}), obtain: {\bf evaluate} $h_2$ {\bf if}:
\begin{equation}
{t_2+t_o}<\frac {p_h \left[{t_c} + b (n)t_1+(b(n)+1){t_o}\right]} {1-p_hb(n)}
\label{eqn:criterion-expanded}
\end{equation}
The factor $\frac {p_h} {1-p_hb(n)}$ depends on the potentially unknown
probability $p_h$, making it difficult to reach the optimum decision.
However, if our goal is just to do better than \lazyastar, then it is safe to replace $p_h$ by an upper bound on $p_h$.
Note that the values $p_h, t_1, t_2, t_c$ may actually be variables that depend in complicated ways on the state of the search.
Despite that, the very crude model we use, assuming that they are setting-specific constants, is sufficient
to achieve improved performance, as shown in Section \ref{sec:empirical}.

We now turn to implementation-specific estimation of the runtimes.
\OPEN~in \astar~is frequently implemented as a priority queue, and thus we have, approximately,
$t_o=\tau \log N_o$ for some $\tau$, where the size of \OPEN~is $N_o$.
Evaluating $h_1$ is cheap in many domains, as is the
case with Manhattan Distance (MD) in discrete domains, $t_o$ is the most significant part of
$t_{e}$. In such cases,
rule (\ref{eqn:criterion-expanded}) can be approximated as~\ref{eqn:criterion-gamma}:
\begin{align}
  &\mbox{\bf evaluate }h_2\mbox{ \bf if } t_2 < \frac {\tau p_h} {1-p_hb(n)} (b(n)+1)\log N_o
\label{eqn:criterion-gamma}
\end{align}
Rule (\ref{eqn:criterion-gamma})
recommends to evaluate $h_2$ mostly at late stages of the search,
when the open list is large, and in nodes with a higher branching factor.

In other domains, such as planning, both $t_1$ and $t_2$ are
significantly greater than both $t_o$ and $t_c$, and terms
not involving $t_1$ or $t_2$ can be dropped from
(\ref{eqn:criterion-expanded}), resulting in Rule (\ref{eqn:criterion-beta}):
\begin{align}
  &\mbox{\bf evaluate }h_2\mbox{ \bf if } \frac{t_2}{t_1} < \frac {p_hb(n)} {1-p_hb(n)}
\label{eqn:criterion-beta}
\end{align}
The right hand side of (\ref{eqn:criterion-beta}) grows with $b(n)$, and here it is beneficial to evaluate $h_2$
only for nodes with a sufficiently large branching factor.


\vspace{-0.1cm}
\section{Empirical evaluation}\label{sec:empirical}
\vspace{-0.1cm}

We now present our empirical evaluation of \lazyastar~and \rationallazyastar, on
variants of the 15-puzzle and on planning domains.

\subsection{Weighted 15 puzzle}

We first provide evaluations on the weighted 15-puzzle variant~\cite{thayer:bss}, where the cost of moving each tile is equal to the number on the tile.
We used a subset of 36 problem instances (out of the 100 instances of~\citet{BFID85}) which could
be solved with 2Gb of RAM and 15 minutes timeout using the Weighted Manhattan
heuristic (WMD) for $h_1$. As the expensive and informative heuristic $h_2$  we use a heuristic based on
lookaheads~\cite{DBLP:conf/aaai/SternKFH10}. Given a bound $d$ we applied a
bounded depth-first search from a node $n$ and backtracked when we reached leaf
nodes $l$ for which $g(l)+WMD(l)> g(n)+WMD(n)+d$. $f$-values from leaves were
propagated to $n$.

Table~\ref{tbl:rational-lazy-a-star} presents the results averaged on all instances solved. The runtimes are reported relative to
the time of \astar~with WMD (with no lookahead), which generated 1,886,397 nodes (not reported in the table). The first 3 columns of Table
\ref{tbl:rational-lazy-a-star} show the results for \astar~with the lookahead
heuristic for different lookahead depths. The best time is achieved
for lookahead 6 (0.588 compared to \astar~with WMD). The fact that the time does not continue to decrease with
deeper lookaheads is clearly due to the fact that although the resulting heuristic improves as a function of
lookahead depth (expanding and generating fewer nodes), the increasing overheads of computing the heuristic eventually outweights savings due to
fewer expansions.

The next 4 columns show the results for \lazyastar~with WMD as $h_1$, lookahead
as $h_2$, for different lookahead depths.  The {\em Good1} column presents the
number of nodes where \lazyastar~saved the computation of $h_2$ while the
$h_2$ column presents the number of nodes where $h_2$ was computed. Roughly
$28\%$ of nodes were {\em Good1} and since $t_2$ was the most dominant time
cost, most of this saving is reflected in the timing results.  The best results
are achieved for lookahead 8, with a runtime of 0.527 compared to \astar~with
WMD.


The final columns show the results of \rationallazyastar~, with the values of $\tau, p_h, t_2$ calibrated for each lookahead depth using a small
subset of problem instances. The {\em Good2} column counts the number of
times that \rationallazyastar~decided to bypass the $h_2$ computation.
Observe that \rationallazyastar~outperforms \lazyastar, which in turn
outperforms \astar, for most lookahead depths.
The lowest time with \rationallazyastar~(0.371 of \astar~with WMD)  was obtained for lookahead 10.
That is achieved as the more expensive $h_2$ heuristic is computed less often, reducing its effective computational overhead,
with some adverse effect in the number of expanded nodes.
Although \lazyastar~expanded fewer nodes, \rationallazyastar~performed much
fewer $h_2$ computations as can be seen in the table, resulting in decreased overall runtimes.

\subsection{Planning domains}

\begin{table*}[t]\vspace{-0.7cm}
\begin{centering}
\tiny{
\begin{tabular}{|l|r|r|r|r|r|r||r|r|r|r|r|r||r|r|r|r|r|r||r|r|}
\hline &
\multicolumn{6}{|c||}{Problems Solved } &
\multicolumn{6}{|c||}{Planning Time (seconds)} &
\multicolumn{2}{|c|}{GOOD } \\
\hline Domain &
$h_{LA}$ & lmcut & max & selmax & \lazyastar& \rationallazyastar &
$h_{LA}$ & lmcut & max & selmax & \lazyastar& \rationallazyastar &
\lazyastar& \rationallazyastar \\

\hline airport & 25 & 24 & 26 & 25 & \textbf{29} & \textbf{29} & \textbf{0.29} & 0.57 & 0.5 & 0.33 & 0.38 & 0.38 & 0.48 & 0.67 \\
barman-opt11 & \textbf{4} & 0 & 0 & 0 & 0 & 3 & N/A & N/A & N/A & N/A & N/A & N/A & N/A & N/A \\
blocks & 26 & 27 & 27 & 27 & \textbf{28} & \textbf{28} & 1.0 & \textbf{0.65} & 0.73 & 0.81 & 0.67 & 0.67 & 0.19 & 0.21 \\
depot & \textbf{7} & 6 & 5 & 5 & 6 & 6 & \textbf{2.27} & 2.69 & 3.17 & 3.14 & 2.73 & 2.75 & 0.06 & 0.06 \\
driverlog & 10 & \textbf{12} & \textbf{12} & \textbf{12} & \textbf{12} & \textbf{12} & 2.65 & \textbf{0.29} & 0.33 & 0.36 & 0.3 & 0.31 & 0.09 & 0.09 \\
elevators-opt08 & 12 & \textbf{18} & 17 & 17 & 17 & 17 & 14.14 & 4.21 & 4.84 & 4.85 & \textbf{3.56} & 3.64 & 0.27 & 0.27 \\
elevators-opt11 & 10 & \textbf{14} & \textbf{14} & \textbf{14} & \textbf{14} & \textbf{14} & 26.97 & 8.03 & 9.28 & 9.28 & \textbf{6.64} & 6.78 & 0.28 & 0.28 \\
floortile-opt11 & 2 & \textbf{6} & \textbf{6} & \textbf{6} & \textbf{6} & \textbf{6} & 8.52 & \textbf{0.44} & 0.6 & 0.58 & 0.5 & 0.52 & 0.02 & 0.02 \\
freecell & \textbf{54} & 10 & 36 & 51 & 41 & 41 & \textbf{0.16} & 7.34 & 0.22 & 0.24 & 0.18 & 0.18 & 0.86 & 0.86 \\
grid & \textbf{2} & \textbf{2} & 1 & \textbf{2} & \textbf{2} & \textbf{2} & \textbf{0.1} & 0.16 & 0.18 & 0.34 & 0.15 & 0.15 & 0.17 & 0.17 \\
gripper & \textbf{7} & 6 & 6 & 6 & 6 & 6 & \textbf{0.84} & 1.53 & 2.24 & 2.2 & 1.78 & 1.25 & 0.01 & 0.4 \\
logistics00 & \textbf{20} & 17 & 16 & \textbf{20} & 19 & 19 & \textbf{0.23} & 0.57 & 0.68 & 0.27 & 0.47 & 0.47 & 0.51 & 0.51 \\
logistics98 & 3 & \textbf{6} & \textbf{6} & \textbf{6} & \textbf{6} & \textbf{6} & 0.72 & \textbf{0.1} & 0.1 & 0.11 & \textbf{0.1} & \textbf{0.1} & 0.07 & 0.07 \\
miconic & \textbf{141} & 140 & 140 & \textbf{141} & \textbf{141} & \textbf{141} & \textbf{0.13} & 0.55 & 0.58 & 0.57 & 0.16 & 0.16 & 0.87 & 0.88 \\
mprime & 16 & 20 & 20 & 20 & \textbf{21} & 20 & 1.27 & 0.5 & 0.51 & 0.5 & \textbf{0.44} & 0.45 & 0.25 & 0.25 \\
mystery & 13 & \textbf{15} & \textbf{15} & \textbf{15} & \textbf{15} & \textbf{15} & 0.71 & \textbf{0.35} & 0.38 & 0.43 & 0.36 & 0.37 & 0.3 & 0.3 \\
nomystery-opt11 & \textbf{18} & 14 & 16 & \textbf{18} & \textbf{18} & \textbf{18} & \textbf{0.18} & 1.29 & 0.58 & 0.25 & 0.33 & 0.33 & 0.72 & 0.72 \\
openstacks-opt08 & 15 & \textbf{16} & 14 & 15 & \textbf{16} & \textbf{16} & 2.88 & \textbf{1.68} & 3.89 & 3.03 & 2.62 & 2.64 & 0.44 & 0.45 \\
openstacks-opt11 & 10 & \textbf{11} & 9 & 10 & \textbf{11} & \textbf{11} & 13.59 & \textbf{6.96} & 19.8 & 14.44 & 12.03 & 12.06 & 0.43 & 0.43 \\
parcprinter-08 & 14 & \textbf{18} & \textbf{18} & \textbf{18} & \textbf{18} & \textbf{18} & 0.92 & \textbf{0.36} & 0.37 & 0.38 & 0.37 & 0.37 & 0.17 & 0.26 \\
parcprinter-opt11 & 10 & \textbf{13} & \textbf{13} & \textbf{13} & \textbf{13} & \textbf{13} & 2.24 & \textbf{0.56} & 0.6 & 0.61 & 0.58 & 0.59 & 0.14 & 0.17 \\
parking-opt11 & 1 & 1 & 1 & \textbf{3} & 2 & 2 & 9.74 & 22.13 & 17.85 & 7.11 & \textbf{6.33} & 6.43 & 0.64 & 0.64 \\
pathways & 4 & \textbf{5} & \textbf{5} & \textbf{5} & \textbf{5} & \textbf{5} & 0.5 & \textbf{0.1} & \textbf{0.1} & \textbf{0.1} & \textbf{0.1} & \textbf{0.1} & 0.1 & 0.12 \\
pegsol-08 & \textbf{27} & \textbf{27} & \textbf{27} & \textbf{27} & \textbf{27} & \textbf{27} & 1.01 & \textbf{0.84} & 1.2 & 1.1 & 1.06 & 0.95 & 0.04 & 0.42 \\
pegsol-opt11 & \textbf{17} & \textbf{17} & \textbf{17} & \textbf{17} & \textbf{17} & \textbf{17} & 4.91 & \textbf{3.63} & 5.85 & 5.15 & 4.87 & 4.22 & 0.04 & 0.38 \\
pipesworld-notankage & \textbf{16} & 15 & 15 & \textbf{16} & 15 & 15 & \textbf{0.5} & 1.48 & 1.12 & 0.85 & 0.9 & 0.91 & 0.42 & 0.42 \\
pipesworld-tankage & \textbf{11} & 8 & 9 & 9 & 9 & 9 & \textbf{0.36} & 2.24 & 1.02 & 0.47 & 0.69 & 0.71 & 0.62 & 0.62 \\
psr-small & \textbf{49} & 48 & 48 & \textbf{49} & 48 & 48 & \textbf{0.15} & 0.2 & 0.21 & 0.19 & 0.19 & 0.18 & 0.17 & 0.49 \\
rovers & 6 & \textbf{7} & \textbf{7} & \textbf{7} & \textbf{7} & \textbf{7} & 0.74 & \textbf{0.41} & 0.45 & 0.45 & 0.41 & 0.42 & 0.47 & 0.47 \\
scanalyzer-08 & 6 & \textbf{13} & \textbf{13} & \textbf{13} & \textbf{13} & \textbf{13} & 0.37 & \textbf{0.25} & 0.27 & 0.27 & 0.26 & 0.26 & 0.06 & 0.06 \\
scanalyzer-opt11 & 3 & \textbf{10} & \textbf{10} & \textbf{10} & \textbf{10} & \textbf{10} & \textbf{0.59} & 0.64 & 0.75 & 0.73 & 0.67 & 0.68 & 0.05 & 0.05 \\
sokoban-opt08 & 23 & 25 & 25 & 24 & 26 & \textbf{27} & 3.94 & 1.76 & 2.19 & 2.96 & 1.9 & \textbf{1.32} & 0.04 & 0.4 \\
sokoban-opt11 & \textbf{19} & \textbf{19} & \textbf{19} & 18 & \textbf{19} & \textbf{19} & 7.26 & 2.83 & 3.66 & 5.19 & 3.1 & \textbf{2.02} & 0.03 & 0.46 \\
storage & 14 & \textbf{15} & 14 & 14 & \textbf{15} & \textbf{15} & \textbf{0.36} & 0.44 & 0.49 & 0.45 & 0.44 & 0.42 & 0.21 & 0.28 \\
tidybot-opt11 & \textbf{14} & 12 & 12 & 12 & 12 & 12 & \textbf{3.03} & 16.32 & 17.55 & 9.35 & 15.67 & 15.02 & 0.11 & 0.18 \\
tpp & \textbf{6} & \textbf{6} & \textbf{6} & \textbf{6} & \textbf{6} & \textbf{6} & 0.39 & \textbf{0.22} & 0.23 & 0.23 & 0.22 & 0.22 & 0.32 & 0.4 \\
transport-opt08 & \textbf{11} & \textbf{11} & \textbf{11} & \textbf{11} & \textbf{11} & \textbf{11} & 1.45 & \textbf{1.24} & 1.41 & 1.54 & 1.25 & 1.26 & 0.04 & 0.04 \\
transport-opt11 & \textbf{6} & \textbf{6} & \textbf{6} & \textbf{6} & \textbf{6} & \textbf{6} & 12.46 & \textbf{8.5} & 10.38 & 11.13 & 8.56 & 8.61 & 0.0 & 0.0 \\
trucks & 7 & \textbf{9} & \textbf{9} & \textbf{9} & \textbf{9} & \textbf{9} & 4.49 & \textbf{1.34} & 1.52 & 1.44 & 1.41 & 1.42 & 0.07 & 0.07 \\
visitall-opt11 & 12 & 10 & \textbf{13} & 12 & \textbf{13} & \textbf{13} & 0.2 & 0.34 & 0.19 & \textbf{0.18} & 0.18 & 0.18 & 0.38 & 0.38 \\
woodworking-opt08 & 12 & \textbf{16} & \textbf{16} & \textbf{16} & \textbf{16} & \textbf{16} & 1.08 & 0.71 & 0.75 & 0.75 & \textbf{0.66} & 0.67 & 0.56 & 0.56 \\
woodworking-opt11 & 7 & \textbf{11} & \textbf{11} & \textbf{11} & \textbf{11} & \textbf{11} & 5.7 & 2.86 & 3.15 & 3.01 & \textbf{2.55} & 2.58 & 0.52 & 0.52 \\
zenotravel & 8 & \textbf{11} & \textbf{11} & \textbf{11} & \textbf{11} & \textbf{11} & 0.38 & \textbf{0.14} & 0.14 & 0.14 & 0.14 & 0.14 & 0.17 & 0.19 \\
\hline
OVERALL & 698 & 697 & 722 & 747 & 747 & \textbf{750} & 1.18 & 0.98 & 0.98 & 0.89 & 0.79 & \textbf{0.77} & 0.27 & 0.34 \\
\hline
\end{tabular}\vspace{-0.2cm}
}
\begin{small}
\caption{\label{t:planning} Planning Domains --- Number of Problems
Solved, Total Planning Time, and Fraction of Good Nodes}\vspace{-0.3cm}
\end{small}
\end{centering}
\end{table*}

We implemented \lazyastar~and \rationallazyastar~on top of the Fast Downward
planning system \cite{helmert:jair-2006}, and experimented with two state
of the art heuristics: the admissible landmarks heuristic $h_{LA}$ (used as $h_1$)
\cite{karpas-domshlak:ijcai-2009}, and the landmark cut heuristic $h_{LMCUT}$
\cite{helmert-domshlak:icaps-2009} (used as $h_2$).
On average, $h_{LMCUT}$ computation is 8.36 times more expensive than $h_{LA}$
computation.
We did not implement HBP in the planning domains as the heuristics we use are
not consistent and in general the operators are not invertible. We also did not
implement OB, as the cost of \OPEN~operations in planning is trivial compared
to the cost of heuristic evaluations.

We experimented with all planning domains without conditional
effects and derived predicates (which the heuristics we used do not support)
from previous IPCs. We compare the performance of \lazyastar~and
\rationallazyastar~to that of \astar~using each of the heuristics
individually, as well as to their max-based combination, and their combination
using selective-max (Sel-MAX)~\cite{domshlak-et-al:jair-2012}. 
The search was limited to 6GB memory, and 5 minutes of CPU time on a
single core of an Intel E8400 CPU with 64-bit Linux OS.

When applying \rationallazyastar~in planning domains we evaluate
rule (\ref{eqn:criterion-beta}) at every state. This rule involves two unknown
quantities: $\frac{t_2}{t_1}$, the ratio between heuristic computations times,
and $p_h$, the probability that $h_2$ is helpful.
Estimating $\frac{t_2}{t_1}$ is quite easy --- we simply use the average computation
times of both heuristics, which we measure as search progresses.

Estimating $p_h$ is not as simple. While it is possible to empirically
determine the best value for $p_h$, as done for the weighted 15 puzzle,
this does not fit the paradigm of domain-independent planning.
Furthermore, planning domains are very different from each other,
and even problem instances in the same domain are of varying size, and thus
getting a single value for $p_h$ which works well for many problems is difficult.
Instead, we vary our estimate of $p_h$ adaptively during search.
To understand this estimate,
first note that if $n$ is a node at which $h_2$ was helpful, then we computed
$h_2$ for $n$, but did not expand $n$.
Thus, we can use the number of states for which we computed $h_2$ that were not
yet expanded (denoted by $A$), divided by the number of states for which we
computed $h_2$ (denoted by $B$), as an approximation of $p_h$.
However, $\frac{A}{B}$ is not likely to be a stable estimate at the beginning of
the search, as $A$ and $B$ are both small numbers.
To overcome this problem, we ``imagine'' we have observed $k$ examples, which
give us an estimate of $p_h = p_{init}$, and use a weighted average between
these $k$ examples, and the observed examples --- that is, we estimate $p_h$ by
$(\frac{A}{B} \cdot B + p_{init} \cdot k) / (B + k)$.
In our empirical evaluation, we used $k=1000$ and $p_{init} = 0.5$.

%
%

Table~\ref{t:planning} depicts the experimental results.
The leftmost part of the table shows the number of solved problems in each
domain. As the table demonstrates, \rationallazyastar~solves the most problems,
and \lazyastar~solves the same number of problems as Sel-MAX. Thus, both
\lazyastar~and \rationallazyastar~are state-of-the-art in cost-optimal planning.
Looking more closely at the results, note that Sel-MAX
solves 10 more problems than \lazyastar~and \rationallazyastar~in the freecell
domain. Freecell is one of only three domains in which $h_{LA}$ is more
informed than $h_{LMCUT}$ ~(the other two are nomystery-opt11 and visitall-opt11), violating the basic assumptions behind
\lazyastar~ that $h_2$ is more informed than $h_1$. If
we ignore these domains, both \lazyastar~and \rationallazyastar~solve more
problems than Sel-MAX.

The middle part of the Table~\ref{t:planning} shows the geometric mean of
planning time in each domain, over the commonly solved problems (i.e., those
that were solved by all 6 methods). \rationallazyastar~is the fastest overall,
with \lazyastar~second. It is important to note that both \lazyastar~and
\rationallazyastar~are very robust, and even in cases where they are not the
best they are never too far from the best. For example, consider the {\em
miconic} domain. Here,  $h_{LA}$ is very informative and thus the variant that
only computed $h_{LA}$ is the best choice (but a bad choice overall). Observe
that both  \lazyastar~and \rationallazyastar~saved 86\% of $h_{LMCUT}$
computations, and were very close to the best algorithm in this extreme case.
In contrast, the other algorithms that consider both heuristics (max and
Sel-MAX) performed very poorly here (more than three times slower).

The rightmost part of Table~\ref{t:planning} shows the average
fraction of nodes for which \lazyastar~ and \rationallazyastar~did not evaluate
the more expensive heuristic, $h_{LMCUT}$, over the problems solved by both
these methods.
This is shown in the {\em good} columns. Our first observation is that this
fraction varies between different domains, indicating why \lazyastar~works well
in some domains, but not in others. Additionally, we can see that in domains
where there is a difference in this number between \lazyastar~and
\rationallazyastar, \rationallazyastar~usually performs better in terms of
time. This indicates that when \rationallazyastar~decides to skip the computation of the expensive heuristic, it is usually the right decision.

\begin{table}[]\vspace{-0.5cm}
\begin{center}
\begin{small}
\begin{tabular}{|l|r|r|}
\hline
&  Expanded & Generated\\
\hline
$h_{LA}$     & 183,320,267 & 1,184,443,684\\
lmcut    & 23,797,219 & 114,315,382\\
\astarmax          & 22,774,804 & 108,132,460\\
selmax          & 54,557,689 & 193,980,693\\
\hline
$\lazyastar$  & 22,790,804 & 108,201,244\\
$\rationallazyastar$   & 25,742,262 & 110,935,698\\
\hline
\end{tabular}
\end{small}
\end{center}\vspace{-0.4cm}
\caption{\label{t:states} Total Number of Expanded and Generated States}\vspace{-0.2cm}
\end{table}

Finally, Table~\ref{t:states} shows the total number of expanded and generated
states over all commonly solved problems. \lazyastar~is indeed as informative
as \astarmax~(the small difference is caused by tie-breaking), while
\rationallazyastar~is a little less informed and expands slightly more nodes.
However, \rationallazyastar~is much more informative than its ``intelligent''
competitor - Sel-MAX, as these are the only two algorithms in our set
which selectively omit some heuristic computations.
\rationallazyastar~generated almost half of the nodes compared to Sel-MAX, suggesting that its decisions are better.


%

\subsection{Limitations of \lazyastar: 15 puzzle example}\label{sec:15puzzle}

Some domains and heuristic settings will not achieve time speedup with L\astar. An example is the regular, unweighed 15-puzzle.
Results for \astarmax~and \lazyastar~with and without HBP on the 15-puzzle
are reported in Table~\ref{tab:3x3runtime}.
$HBP_1$ ($HBP_2$)  count the number of nodes where HBP pruned the need to compute $h_1$ (resp. $h_2$).
OB is the number of nodes where OB was helpful. {\em Bad} is the number of nodes
that went through two \OPEN~cycles.
Finally, {\em Good} is the number of nodes where computation of $h_2$ was saved due to \lazyastar.

In the first experiment, Manhattan distance (MD) was divided into
two heuristics: $\Delta x$ and $\Delta y$ used as $h_1$ and $h_2$.
Results are averaged over 100 random instances with average solution
depth of 26.66. As seen from the first two lines, HBP when applied on top of \astarmax
saved about 36\% of the heuristic evaluations.
Next are results for  \lazyastar~and \lazyastar+HBP. Many nodes are pruned by HBP, or OB.
The number of {\em good} nodes dropped from 28\% (Line 3) to as little as 11\% when HBP was applied.
Timing results (in ms) show that all variants performed equally.
The reason is that the  time overhead of the $\Delta x$ and $\Delta y$
heuristics is very small so the saving on these 28\% or 11\% of nodes was not
significant to outweigh the HBP overhead of handling the upper and lower bounds.

The next experiment is with MD as $h_1$ and a variant of the
additive 7-8 PDBs~\cite{ADBAIJ02}, as $h_2$. Here we can observe an interesting phenomenon. For \lazyastar, most nodes were caught by either
HBP (when applicable) or by OB. Only 4\% of the nodes were {\em good} nodes.
The reason is that the 7-8 PDB heuristic always {\em dominates} MD and is
always the maximum among the two. Thus, 7-8 PDB was needed at early stages (e.g. by OB) and MD itself almost
never caused nodes to be added to \OPEN~and remain there until the goal was found.

These results indicate that on such domains, \lazyastar~has limited merit. Due to uniform operator cost and the heuristics being consistent and simple to compute, very little space is left for improvement with {\em good} nodes. We thus conclude that \lazyastar~is likely to be effective when there is significant difference between $t_1$ and $t_2$, and/or operators that are not bidirectional and/or with non-uniform costs, allowing for more {\em good} nodes and significant time saving.

\begin{table}\vspace{-0.5cm}
    \begin{centering}
     \tiny{
    \begin{tabular}{|l|r|r|r|r|r|r|r|}

	\hline
	Alg. & Generated &  HBP1 & HBP2 & OB & Bad & Good & time \\
    \hline
    \multicolumn{8}{|c|}{$h1 =\Delta X$, $h2 =\Delta Y$, Depth = 26.66}\\
		\hline
		A* & 1,085,156 & 0 & 0 & 0 &  0 & 0 & {\bf 415} \\
		A*+HBP & 1,085,156 & 216,689 & 346,335 & 0  & 0 & 0 & 417 \\
		LA* & 1,085,157 & 0 & 0 & 734,713 & 37,750 & 312,694 & 417 \\
		LA*+HBP & 1,085,157 & 140,746 & 342,178 & 589,893& 37,725 & 115,361 & 416 \\
		\hline
      \multicolumn{8}{|c|}{$h1 =$ Manhattan distance, $h2 =$ 7-8 PDB, Depth 52.52}\\
		\hline
		A* & 43,741 & 0 & 0 & 0 & 0 & 0 & 34.7 \\
		A*+HBP & 43,804 & 30,136 & 1,285 & 0 & 0 & 0 & 33.6 \\
		LA* & 43,743 & 0 & 0 & 42,679  & 47 & 1,017 & 34.2 \\
		LA*+HBP& 43,813 & 7,669 & 1,278 & 42,271  & 21 & 243 & {\bf 33.3} \\
		\hline
   \end{tabular}}
   \vspace{-0.3cm}
	\caption{Results on the 15 puzzle}\vspace{-0.2cm}
	\label{tab:3x3runtime}
\end{centering}
\end{table}

\vspace{-0.1cm}
\section{Conclusion}\vspace{-0.1cm}

We discussed two schemes for decreasing heuristic evaluation times. \lazyastar~is very simple to implement
and is as informative as \astarmax. \lazyastar~can significantly speed up the search, especially if $t_2$ dominates
the other time costs, as seen in weighted 15 puzzle and planning domains.
Rational \lazyastar~allows additional cuts in $h_2$ evaluations, at the expense
of being less informed than \astarmax. However, due to a rational tradeoff, this
allows for an additional speedup, and Rational \lazyastar~achieves the best
overall performance in our domains.

\rationallazyastar~is simpler to implement than its direct
competitor, Sel-MAX, but its decision can be more informed.
When \rationallazyastar~has to decide whether to compute $h_2$ for some node
$n$, it already {\em knows} that $f_1(n) \leq \optcost$.
By contrast, although Sel-MAX uses a much more complicated decision
rule, it makes its decision when $n$ is first generated, and does not know whether $h_1$ will be informative enough to
prune $n$. Rational \lazyastar~outperforms Sel-MAX in our planning
experiments.

\rationallazyastar~and its analysis can be seen as an instance
of the rational meta-reasoning framework~\cite{RussellWefald}. While this framework
is very general, it is extremely hard to apply in practice. Recent work
exists on meta-reasoning in DFS algorithms for CSP) \cite{DBLP:conf/ijcai/TolpinS11} and in Monte-Carlo tree search~\cite{DBLP:conf/uai/HayRTS12}. This paper applies
these methods successfully to a variant of~\astar.
There are numerous other ways to use rational meta-reasoning to improve~\astar, starting from generalizing
\rationallazyastar~to handle more than two heuristics, to using the meta-level to control
decisions in other variants of \astar. All these potential extensions provide fruitful ground for future work.

\vspace{-0.1cm}
\section{Acknowledgments}\vspace{-0.1cm}

The research was supported by the Israeli Science Foundation (ISF) under grant \#305/09 to Ariel Felner and Eyal Shimony
and by Lynne and William Frankel Center for Computer Science.

\bibliography{paper}
\bibliographystyle{named}

\end{document}